# New Ideas for Brain Modelling

Kieran Greer[1]

[1]*Distributed Computing Systems, Belfast, UK.*

***Abstract: -*** This paper describes some biologically-inspired processes that could be used to build the sort of networks that we associate with the human brain. New to this paper, a 'refined' neuron will be proposed. This is a group of neurons that by joining together can produce a more analogue system, but with the same level of control and reliability that a binary neuron would have. With this new structure, it will be possible to think of an essentially binary system in terms of a more variable set of values. The paper also shows how recent research associated with the new model, can be combined with established theories, to produce a more complete picture. The propositions are largely in line with conventional thinking, but possibly with one or two more radical suggestions. An earlier cognitive model can be filled in with more specific details, based on the new research results, where the components appear to fit together almost seamlessly. The intention of the research has been to describe plausible 'mechanical' processes that can produce the appropriate brain structures and mechanisms, but that could be used without the magical 'intelligence' part that is still not fully understood. There are also some important updates from an earlier version of this paper.

## I.   INTRODUCTION

This paper describes some biologically-inspired processes that could be used to build the sort of networks that we associate with the human brain. The biological background is probably only what an AI researcher, interested in the area, would know about; but the proposed processes might offer a new way of looking at the problem. So while a computer model or simulator is the goal, the processes map more closely to a real human brain, at least in theory. This paper firstly considers how a more refined processing unit can be constructed from similar binary-operating components. If studying this topic, it is easy to think in binary terms only, but the question of how our sophisticated intelligence derives from that is difficult to answer. The thought process might try to find a particularly clever or complicated function that works inside of a neuron, to compensate for its apparent simplicity. The computer scientist would also try to use a distributed architecture, to combine several units for an operation, again to make the whole process more complex and variable. That is the direction of this paper, but it will attempt to show that the modeller's thinking process can begin to look at a collection of neurons as a single unit that can deliver a complex and variable set of values reliably. It can be constructed relatively easily and with some degree of accuracy, by a mechanical process that the simple neuron might not be thought to accommodate.

The second half of the paper gives a detailed review of related work. Some of this is quite old now, but it helps to show that the basic principles for building these neural systems are included. This paper then recaps on some other recent research and is able to demonstrate a new level of detail that integrates the new technology almost seamlessly. A completely new model is not proposed, where the 3-level architecture described in [8] or [11], is still being looked at. That model contains a number of quite abstract components and the recent research can replace those with more detailed ones. That is, they would be able to fit in somewhere, even if the whole picture is not complete. Some mathematical theories help to back this up, and even if the proposed solutions are not 100% biologically proven, they would be useful for building a computer-generated system. The idea and importance of resonance is outlined and proposed as part of a new theory, even if it is not defined exactly in the current model.

The rest of this paper is structured as follows: section II describes some background work that can help to explain the reasons for a new model. Section III describes the refined neuronal component. Section IV gives an alternative feedback mechanism that is essential for the new model. It also provides an example set of equations that might be used to simulate the system. This is only an example set, to show how relatively simple it would be to re-produce the desired activity. Section V describes a plausible process for linking these neurons automatically and in an arbitrary way. Section VI extends section II, listing other related work. Section VII puts the new neuron and a lot of the earlier research into context and section VIII adds more detail to the current cognitive model. Finally, section IX gives some comments on the work.

## II. BACKGROUND TO THE PROPOSAL

A lot of research on this topic has described how neurons and synapses can grow and cluster to form more complex structures. There are also different types of entity with slightly different functionality, but it is still very unclear what exactly goes on and it is difficult to map the processes onto basic computer models. The superficially simplistic nature of neurons is written about in [26] for example, which discusses that a binary signal is limiting. It then explains that adding more states to a single unit comes at a price and that it has been worked out mathematically that the binary unit with 2 states is the most efficient. This is because the signal also needs to be decoded, which is expensive. Therefore, if more states can be achieved through different connection sets instead, this would be desirable and it might also add new knowledge to the network. Also in [26] chapter 2, it is explained how any logic function can be built from the McCulloch-Pitts neurons [22] and also a hint at the new architecture in this paper, by pointing out that any weighted neuron can be replaced by an unweighted one, with additional 'redundant' connections. A weight value of 3, for example, can be replaced with a single connection that is split into 3 inputs, but that is definitely a different design. The author has listed a number of papers ([6]-[12]) that were investigated separately, but might still provide a coherent view over all of them. They have made lots of suggestions that also turn out to be mostly established theories. For example, hierarchies, reasons for duplicating, use of patterns, timing and how to cluster; which are also of course, the established theories. There are also one or two more radical ideas however, including this paper, which could change the failure of a previous theory.

With regard to automatic network construction, there has been a lot of research looking at modelling the brain neuron more closely, including binary units with thresholds and excitatory or inhibitory inputs. The paper [28] is one example that focuses on dynamic stimulus-driven models. It describes one equation for the firing rate that includes external sensory inputs, as well as from other neurons. It also states that for the firing to be sustained, that is, to be relevant, requires sufficient feedback from other firing neurons. Once the process starts however, this can then excite and bring in more firing neurons, when inhibitory inputs also need to fire, to stop the process from saturating. They note that the network must be sensitive to external stimuli, but useful forms of signal propagation must also be generated internally. The paper [16] shows evidence of a chemospecific steering and aligning process for synaptic connections. That is, there may be some purpose behind how synapses grow and connect. It describes however that this is only the case for certain types of local connection, although other researchers have produced slightly different theories there. The paper [2] gives an argument that a mechanical process, more than an electro-chemical one, is responsible for moving the signals around the brain. This includes mechanical pressure to generate impulses. In [26], only an Ion pump is described, where the activity results solely from Ions moving through a fluid environment. It is argued in [2] however that this is not sufficient to explain the forces involved and some form of mechanical process, controlling pressure, is also required. This can also help linking appendages to grow, as it can influence fluid movement.

A hierarchical network structure for the human brain is well known and is written about in many texts. In the hierarchy, some hidden or middle-layer neurons do not represent anything specifically, but are more for structural or clustering purposes. In [14] it is also noted that there are abstract structures or points in a hierarchy, representing largely how the structure has formed. This can also include lateral connections between hierarchies. There are also 'higher' and 'lower' levels of the hierarchy, as more basic concepts are grouped into higher-level ones. It is then explained that feedback is more common than signals travelling forwards only. This feedback helps to make predictions, or to confirm a certain pattern of firing, as part of an auto-associative 'memory-prediction' structure. The brain needs to remember the past to predict what will happen in the present, to allow you to reason about that. This paper is more concerned with mechanical processes that can operate automatically and together, than the processes that are the intelligent reasoning. It can therefore ignore a very complex model of the neuron itself and consider how to model the pattern-forming mechanisms and the interactions between them.

### III. REFINED NEURONAL UNIT

To explain the theory, a simplified model will be used, with neurons connected only by synapses. The term synapse will cover all of the connectors – including dendrites or axons, for example. Consider a single neuron with a number of inputs from other neurons and a single output. That neuron will be called the 'main' neuron here. Each input transmits a signal with a strength value of 1. The activation function is stepwise, where if the threshold value is met, the neuron outputs a signal with a strength value of 1 and if not, it outputs nothing, or 0. This is essentially a binary function and is being called a 'course' neuron here. All neurons are therefore the same and operate in the same simplistic way. Consider structures that then use these course neurons, as shown in Figure 1. The most basic is a single main neuron with a number of inputs, 5 for example and its own output (1a). If the threshold value is 4, then 4 of the 5 inputs must fire for the neuron itself to fire. Consider a second scenario where the neuron has 25 other neurons that wish to connect with it through their related synapses (1b). In this case, if all of the input neurons connect directly, then only 4 of the 25 neurons need to fire to trigger the main neuron. This is a very low number of inputs and might even cause the concept that the neuron represents to lose most of its meaning.

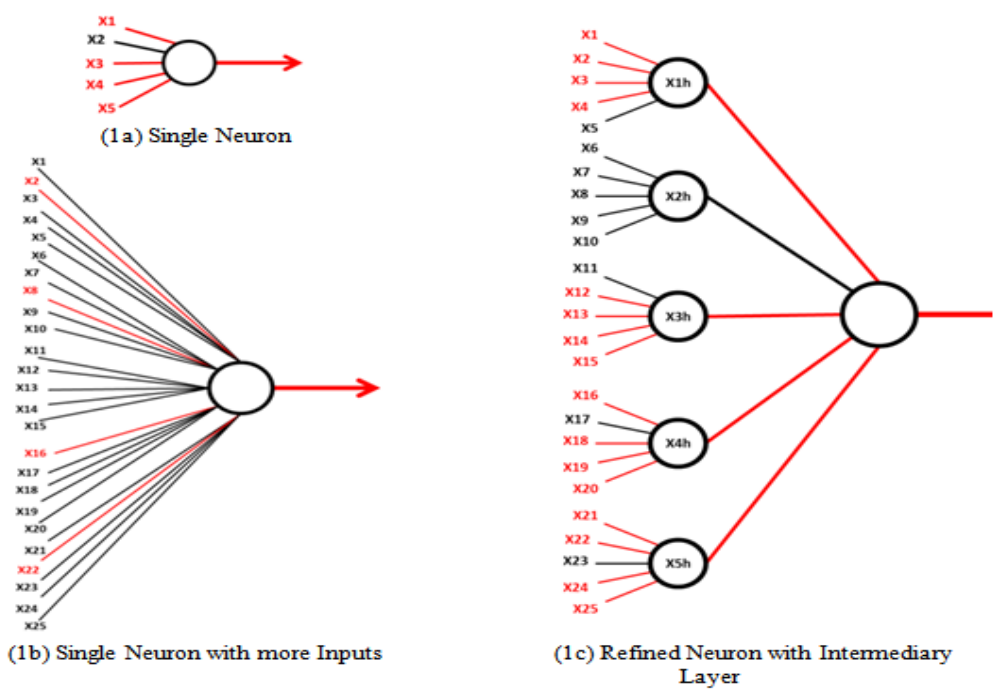

Figure 1. From Course to Refined Neuronal Unit Structure.

What if there are now a number of units as part of a middle or intermediary layer, still made from the same neuronal unit (1c). Maybe 5 of each of the input neurons connect to 1 middle layer neuron. The 5 middle layer neurons then connect with the main neuron. Then, 4 of the 5 intermediary neurons must fire and each of those will only fire if 4 of the related 5 input neurons fire. Therefore, for the main concept neuron to fire now requires 16 of the original input neuron set to fire, which is more meaningful in itself. It is also interesting with relation to the earlier cognitive model that was suggested ([8][11]). This new neuron could operate in the middle aggregating and averaging level of that model. The level below optimises a linking process, while the level above attempts more intelligent interactions. The middle level takes link sets and through some very simple operations can derive a low level of knowledge, such as sum and average a set of links on a particular path. The model was designed with autonomous query processes also in mind [12], which is why it includes this aggregating idea. Processing a larger number of neurons does add stability, because once set, it is more difficult for it to change significantly. It might also trend towards average calculations over specifics, because a greater variety can produce the same result.

Adding intermediary units has also made the structure more analogue, as each input neuron is now represented by a fraction of its true signal strength. In effect, each input neuron now sends a signal represented by one quarter of its true value. This new entity will be called a 'refined' neuron (ReN). The term does not relate specifically to a single neuronal component or to a specific set of them. The term relates, slightly abstractly, to the main neuron plus its set of would-be inputs that have now been converted into a more analogue set of values. The inputs could have derived from any number of other neuron sources. The process can be further modified by allowing some neurons to connect directly, while others go through one or more intermediary layers. Therefore, each input can have a very different weight or influence over the result of another neuron firing and it can all be controlled using the same simple units, but with a more complex set of connections.

## IV. ALTERNATIVE FEEDBACK MECHANISM

While a hierarchical structure is included in most research work on this topic and therefore the ReN architecture must also be implicit, it is difficult to recall a paper that suggests looking at a group of neurons in this way. The importance shifts more to the synaptic connection and structures, with the operation of the neuron itself regarded as more simplistic and constant. The paper [22] notes that synapses can vary in size and transmission speeds and even that the threshold of a neuron can be changed. This could lead to different firing rates for the neuron as well, but for a reliable system that works over and over again, the simplest configuration is the best. This would be to change the relatively static synaptic structure and have the neurons firing, mostly under one state or condition, without additional 'moving parts' that might be prone to damage over time. In [14] it is described how feedback to the same region is more common than signals in one direction only. This is realised by sending a neuron's output signal, back into the input areas again. The idea being that this confirms what the correct patterns are, by completing a circuit and therefore producing reinforcement. This paper will suggest a slightly different form of feedback, but one that must work in conjunction with the cyclic feedback. Instead of completing circuits or cycles, a firing event results in information flowing back up the same synapse channel that it originated from. As will be suggested in later sections, there are in fact some practical advantages to this form of feedback. A repulsive force might simply be the rejection of the Ion channel, when there is too much input, which is consistent with Figure 1. This theory therefore prefers a more physical movement, related to pressure changes.

### A. Signal Behaviour after Firing

The following scenario describes how the input signal might behave: Before firing, the neuron acts like a capacitor and soaks up the input signal until the threshold value is met. As soon as it is triggered, the firing process forces a signal out through any allowed channel. The signal is of course directed mainly through the output channel, but it would be desirable if it could at least block the direction that new input signals would come from. The idea is that instead of reinforcement through completing circuits, after firing, a neuron rejects or blocks the input signal while it recovers. This could cause turbulence in the input flow that might get recognised. Each input channel can become excited from a repeating block. The purpose of the turbulence is that it marks a change from the normal flow and would happen less frequently during the random search. The other option is to speed up the firing rate through a cyclic reinforcement, where this continues until inhibitors switch it off. The inhibitors however use a form of saturation and so to recognise something specific, the feedback must be local. While a general area could significantly change that way, resonance must be key for a more specific type of recognition. One might also think that speeding up the firing rate would add to or even be the cause of the potential turbulence. The model might then want to create new channels to relieve the pressure.

### B.     Feedback Equation Set

This theory is intended for a computer simulation and it should be possible to create some mathematical equations to model the main process relatively easily. For example, to calculate an amount that might get rejected by a neuron can be as simple as:

Let $N$ be the number of direct input synapses from other neurons.
Let $T_m$ be the threshold for the main neuron.
Let $I_s$ be the input signal from a single direct synapse.
Let $I_{sn}$ be the total input signal from all direct synapses.
Let $AE_{sn}$ be the average excess input value, for each input synapse.
Let $\delta$ be the distance from the firing neuron, with relation to the feedback signal.
Let $AE_{sn\delta}$ be the average excess input value, for each input synapse, at distance $\delta$ from the point of repulsion.

Then:
$$I_{sn} = \sum_N I_s$$
$$AE_{sn} = (I_{sn} - T_m) / N.$$

Each input synapse can be assigned the default value of 1, for example. The main neuron threshold can also be assigned some value, set it to 5 in this case. Therefore after the main neuron itself fires and collapses, rejecting any other signals during that time, it will reject for each synapse the calculated $AE_{sn}$ value. This looks OK mathematically, as it is proportional to the total number of inputs $N$, where more inputs will give a larger average rejection value. It is easy to imagine that as the rejection value becomes larger, it will counter any signal flowing forwards with a force that might cause a disturbance and possible subsequent events. For example, if there are 10 inputs, then the excess is $(10 - 5) / 10 = 0.5$, but if there are 50 inputs, then the excess is $(50 - 5) / 50 = 0.9$.

This calculation is therefore very easy, but a more complete system would need to include other factors, more specifically with modelling the synapse, how much excess signal a neuron absorbs firing and also how the continuing input signal contributes to the resulting rejection force and turbulence. It should also be noted that a repulsive force will also reduce with distance and so a more accurate measurement of it would need to consider the distance it might travel backwards and also the opposing forces that it might meet. Another consideration could be that the forward moving force is weakened by both the repulsive force and any previously blocked input. There are physical theories (thermodynamics) that cover this sort of fluid interaction however. The simplest case is therefore probably inadequate, but can be formulated as: if there is always the same forward force, the resistance to the repulsive force is constant and refreshed, and so it can be mostly measured using distance and static force amounts, as follows:

$$AE_{sn\delta} = AE_{sn} - (\delta \times combined\ forward\ forces).$$

As the stopping criterion is known – a balanced state, it would be possible to test randomly, different sets of values, representing different neuron firing configurations or inputs, and measure time and amounts before the whole system does reach a balanced state. The whole system is still relatively simplistic for modelling.

### V.     AUTOMATIC NETWORK CONSTRUCTION

This section describes a plausible automatic process for constructing a network based on the new neuronal model. It is from a computer science perspective and will suggest mechanical processes that can be simulated in a computer program.

### A. Mechanical Network Process

There is a main neuron with a number of synapse inputs from other similar neuronal units. The inputs send signals, where the signal strength collectively is too strong. The main neuron's threshold is met and so it fires and then simulates the refractory period, by rejecting the excess input that it receives. This excess rejected signal counters more input travelling to the main neuron, where a set of equations relating to section IV.B, determines the strength. Then something like the following might happen:

- This force causes turbulence and stimulates a 'sideways growth' of some sort to the synapse.
- The new synapse growths can meet and join.
- They can also form a new neuron, when only their combined strengths will then produce the 'unit' output value again.
- The new intermediary neuron then links with the main neuron again and the original connection paths can be closed (or partially closed?).
- Note that time is a critical element in the whole process as well, when only neurons firing at the same time will affect each other.
- The new neuron layers would lead to a more balanced system that would not upset the overall energy that flows. They would also naturally start to represent (sub)concepts in their own right, adding new information to the brain.

The schematic of Figure 2 illustrates how an intermediary neuron might form. In figure 2(a), there are 3 neurons sending input signals. The red-brown wavy line represents additional turbulence created when there is too much input and some is ejected backwards. The red-brown semi-circles are areas of the synapse that have been influenced by the energy and start to grow. Two of them meet in figure 2(b) and also a new neuron somehow forms. This grows forward through an attractive stimulus, where it finds the main neuron again. Therefore the input from two of the original neurons is reduced to only one input signal through the new intermediary one. In the figure 2(b), in fact, one of the original input paths has closed completely, while the other is still open, but with a reduced signal. This is only a theory of course.

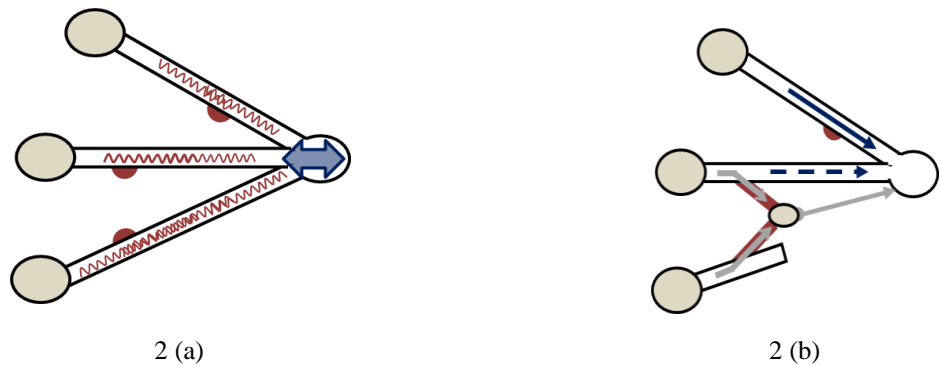

2 (a)                                   2 (b)

Figure 2.Schematic showing the new Intermediary Neuron formation.

### B. Synchronisation Problems

There are one or two obvious synchronisation problems with this mechanical process. The first problem is: two neurons A and B, trying to join but firing at different times, when neuron B fires more often than the group. For example:
- Neuron A fires only with the neuron group that produces the excess input.
- Neuron B also fires during this time, but also at other times.

- When these join, the new intermediary neuron will fire only when both neurons A and B fire, which is OK for neuron A, but not for neuron B.
- Neuron A is part of the excess signal group only and so the main neuron that it connects to needs the re-balancing to adjust for this excess energy. This new intermediary unit does not change the signal that the main neuron receives or when it receives it.
- Neuron B however sends signals at other times to the main neuron. This still needs to be sent, even when the new intermediary neuron does not fire.

One answer to this problem is to make the intermediary layer construction primarily time-dependent and influences the closing of the original paths:
- If the input signal's force is strongly countered, the rejected signal collides and causes turbulence or stagnation. If there is an easier route, this might help the path to close.
- If the measurement requires the amount of rejection to be proportional to the amount of closure; then neurons that only fire as part of an excess set, will always receive the repulsive force and close together. Neurons also firing at other times however will be able to complete the original circuit, which might help to keep the path open.

Another factor here could be sets of competing neurons. If they require slightly different movement directions, then a neuron that fires in sync with several groups can get confused, which could delay it joining with any specific group. There is proof that synapses can shrink or enlarge. So it is only when there is too much input and the system becomes unbalanced, that the turbulence becomes enough to force other types of activity.

There is also the other extreme where neuron B fires less often than the excess group, but the excess group always fires when it does. In that case:
- Its feedback will occur less often and the excess group will have grown new links more quickly, possibly relieving the additional pressure before the less frequently firing neuron is encouraged to complete the new circuit.

### C. Biological Network Comparison

Firstly, to summarise the newly discussed ideas:
1. The system will try to self-organise itself through more intermediary units.
2. The intermediary units create a more balanced system, add meaning to the network and can produce a more varied signal.
3. Neurons firing together can represent a common concept or goal.
4. Synapses firing at the same time can try to link with each other, or combine for related neurons.
5. Resonance is important, where out-of-sync neurons will be able to perform differently.

There are real biological processes that this model can be mapped to. The main neuron firing and then collapsing, when a refractory period prevents it from firing again, is the real biological process. This can help to prevent cycles, or attractors [30], from occurring. If the synapses are still sending input signals, maybe the neuron cannot absorb the signal then and will instead simply block the new input. Inhibitors are typically required to control the firing process, so the process must produce excess signal in any case. In [26] it is explained that selective channels need to be opened and can be closed after depolarisation (firing), but this is for electrical Ion signals more than fluid ones. The Ion flow can still be blocked and influenced using this process however. The paper [2] argues for a mechanical process, with a pressure loss. If that is correct, then a physical block could take place when maybe the signal cannot continue. It is unclear if a fluid signal would be forcibly ejected backwards, which would require some kind of reaction with the synapse membrane. The paper [29] describes that more than one type of sodium channel can be created and that they interact with each other, producing a variable chemical signal. Small currents are also involved, so blocking of 'something' looks possible. If the rejected signal meets a signal travelling in the other direction, this turbulence might stimulate growth in that area. This growth would necessarily be sideways and would help to relieve pressure or excitation. The rest of the process is then clear again, where some of the tasks are required, even without this new theory. Neurons need to form and synapses need to find other neurons, as part of any brain construction. Hebb's law [15] 'neurons that wire together fire together', would also suggest that the closer neurons would be more likely to combine into a single unit, while the paper [16] suggests that an attracting stimulus can control growth direction. So there is a clear set of biological processes that could, at least in theory, map to the mechanical ones.

## VI. RELATED WORK

The related work is divided into sections, supporting different aspects of the new view that is described in section VII. The earlier papers ([6]-[11]) might need to be read at some stage, but a general understanding should be possible without reading them first. Specifically, this paper is more biologically-oriented than might be usual. The author is not particularly expert in that area, but the underlying mechanisms need to be compared with the real biological ones. Other neural network models in particular, are described in detail. The relevance of the book by Hawkins and Blakeslee [14] has already been noted in this and other papers.

### A. Neural Networks

There are lots of other neural network models and it would not be possible to mention all potentially relevant ones, suggesting that the model(s) of this paper contain(s) some popular features. The classic paper by McCulloch and Pitts [22] for example, gives original research on looking at the human brain as a collection of binary operating neurons. They describe a set of theorems under which these neurons behave, including synaptic links with excitatory or inhibitory connections. They argue that if there is a definite set of rules or theorems under which the neurons behave, this can then be described by propositional logic. Early AI favoured logic and knowledge-based systems. A layered or hierarchical structure is implicit in all of the descriptions and used to explain how a temporal factor can be added and possibly measured, through a spatial structure. They then focus more on behaviour and states, rather than precise mathematical values. So while they show how the neurons might fire, the idea of controlling and measuring the signal value accurately, is not addressed as much. Hebb [15] combined up-to-date data about behaviour and the mind into a single theory. Hebb's law is often paraphrased as 'Neurons that fire together wire together'. Positive feedback increases the bond between the firing cells and creates 'attractors'. A combination of his work finally brought together the two realms of human perception that for a long time could not be connected properly. That is, it connected the biological function of the brain as an organ together with the higher function of the mind, but the exact mechanism for this is still unknown. This paper proposes automatic mechanisms for a lower level of function, but does overlap into some higher-level functionality.

Ultimately, resonance in the excited brain area will be the key to the signal being significant. If the search path finds node groups that can agree on the current state and fire together, causing a stronger signal in some way, then this is what the brain will recognise as positive feedback to a search process. Adaptive Resonance Theory [4][13] is an example of trying to use resonance, created by a matching agreement, as part of a neural network model. The primary intuition behind the ART model is that object identification and recognition generally occur as a result of the interaction of 'top-down' observer expectations with 'bottom-up' sensory information. The model postulates that 'top-down' expectations take the form of a memory template or prototype that is then compared with the actual features of an object, as detected by the senses. This comparison gives rise to a matching process with existing categories and produces a measure of difference. As long as this difference does not exceed a set threshold called the 'vigilance parameter', the sensed object will be considered a member of the expected class[1]. The learning process has two stages. An input pattern is matched to existing categories, where the process is self-organising and only one category can fire. Only if the input matches a category, does a training stage take place, to update the related weight values. If there is no match, then an uncommitted neuron is committed and adjusted towards matching the input vector. There are examples of the real human brain using ART methods in some processes [13]. In a theoretical sense, the final model of Figure 6 is quite similar. For that, the bottom-up sensory information is static knowledge, while the top-down observations are dynamic manipulations of the static knowledge.

---

[1] Wikipedia description, http://en.wikipedia.org/wiki/Adaptive_resonance_theory.

Two other models that are related to Hebbian reinforcement and energy equilibrium states are Hopfield neural networks [18] and the stochastic version called Boltzmann machines [1].These are recurrent neural networks that can also act as (auto)associative memories. Like human associations through similarity, they can retrieve whole patterns when presented with only part of the pattern as input. During training, the only goal is a reduction in the global energy state. Neurons learn the underlying statistical characteristics instead of direct adjustments for each pattern. A majority rule can then determine the resulting output. There is evidence that brain activity during sleep employs a Boltzmann-like learning algorithm, in order to integrate new information and memories into its structure [32]. The article also states that 'Neuroscientists have long known that sleep plays an important role in memory consolidation, helping to integrate newly learned information.' The middle layer of the cognitive model in [8] or [11] might benefit from a Boltzmann-like machine, although, it would be against the ReN model suggested here, which would also operate in that aggregating layer. But the idea of automatically balancing and reducing the energy state is the same. One point is that the recurrent neural networks have some problems with storage capacity.

There are also examples of other relevant models. If thinking about equilibrium, then other self-organising neural networks include [5] or [20]. GasNet [27] neural networks use a method of influencing your surrounding area, even if there is no direct synaptic connection. The neuron emits a 'gas' that could be compared to an Ion attraction and it is known to speed up convergence. The paper [5] tries to create a visual pattern recogniser. It is interesting because it tries to copy the brain structure quite closely and has some apparent similarities with the ideas of this paper. That model is self-organising and hierarchical. It also proposes some form of resonance through the backward motion of signals, and includes static and selective learning processes. The problem of noisy input is tackled by allowing a 'region' to represent the same thing, instead of a specific point. This is included as part of the learning process and is implemented through layers that can recognise the same pattern point, but at different positions. The theory behind this is described in [3], who describes different ways of implementing it. Note that this does not add understanding and so is mostly statistical, where different images of the same thing would not necessarily be recognised. There is also an implementation of inhibitory units that can suppress a feature if it is not correct. The pattern recognition occurs when both the forward hierarchical categorisation and the backward reinforcement or resonance agree, to produce the resulting output.

The papers [1] and [3] both note that even connectionist systems require sequential and logical processing to successfully model the human brain. This is also written about in [8]. As illustrated by the previous section, as well as the more popular statistical networks, other models try to copy the human brain more closely. The mathematical theories of some of these are written about in [3], which could act as an introductory tutorial. The section III on Statistical Neurodynamics might be particularly useful, for example. As part of one equation set, there is a clearly defined explanation for an end or terminating node that is part of an arbitrarily connected graph. Nodes can therefore be defined individually, or in arbitrarily linked collections, with states defined between either type. The next section B of this review also describes why terminating nodes are important.

One final group of networks that might be worth mentioning are the evolutionary or hybrid ones. They use the traditional statistical methods, but allow changes through other evolutionary methods, such as genetic algorithms [17]. The paper [21] proposes a biologically plausible mechanism at a single neuron level, for compensating for neural transmission delay. This is the time for the signal to travel in the sensor or brain, during which the real environment might have changed. To cope with this, the brain is required to extrapolate information, to align precisely its internal perceptual state with the real environmental state. They propose to use recurrent neural networks, because the feedback loops make available the history of previous activations (also [14]). The evolution that takes place is an Enforced Subpopulation algorithm (ESP) type. Instead of full networks, single neurons are evolved so that best neurons from each subpopulation can be put together to form a complete network. The network can then learn independent sub-functions and they also give some time-based equations for the activation level of neurons at the synapse connection points. The paper [25] is interesting, because it takes a similarly flexible approach to the one that this paper is suggesting. It proposes to use neural gas networks, where the network connection topology can be learned by a localised stigmergic ant-based algorithm, instead of a global one. They even have the idea of creating intermediary units between influenced areas, to reduce the network error. 'The squared error of the nearest unit in turn is accumulated, so that after λ input signals have been learned, a new unit is created half between the two neighbouring units with the highest accumulated errors.' So this includes the ideas of Hebbian influence, arbitrary structure and the addition of close units to reduce the error.

### B. Minsky's Frames

One of the earlier models by Minsky [24] describes a 'Frame' structure for representing objects or concepts and can be made from networks of nodes with relations. Frames appear to be more logic-based and therefore fit better with semantics and formal knowledge structures. It is interesting here because it also contains the notion of terminals or slots that must be filled with specific instances of data. These terminals are more fixed and can be linked to by any groups of the frame nodes. Therefore, there are different paths to the terminals and a crossing-over, or merging, between paths would be likely. The fame-based system deals strongly with logic and knowledge, and would be intended more for higher-level thought processes, than lower-level pattern recognition. While frames are logic-based, the paper does note the requirement for some form of replacement, possibly analogous to a merging or aggregating operation. When a proposed frame cannot be made to fit reality, the network provides a replacement frame. This inter-frame structure makes it possible to represent other knowledge about facts, analogies, and other information useful in understanding. A search through the linked frames or nodes then also involves a matching process, ending with a match on the terminals. So frames can be compared with pattern ensembles and hierarchies.

### C. Human Biology

This section is more biologically-oriented. The author is not very expert there and the papers are more of a random selection that might make some relevant points. The paper [5] notes that analog signals can also be sent by using different firing frequencies. The paper [28] is one example that gives an equation for the firing rate that includes external sensory inputs as well as input from other neurons. It also states that for the firing to be sustained, that is, to be relevant, requires sufficient feedback from the firing neurons, to maintain the level of excitation. Once the process starts however, this can then excite and bring in other firing neurons, when inhibitory inputs also need to fire, to stop the process from saturating. A weighted equation is given to describe how the process can self-stabilise if 'enough' inhibitory inputs fire. The paper [16] studies the real biological brain and in particular, the chemospecific steering and aligning process for synaptic connections. It notes that there are different types of neuron, synaptic growth and also theories about the processes. Current theory suggests that growth is driven by the neuron itself and also suggests that the neuron is required first, before the synapses can grow to it. This paper requires a charged signal to attract, but it might like a neuron to grow at the end of synapse connections. However, they do note a pairwise chemospecific signalling process, as opposed to something that is just random and they also note that their result is consistent with the known preferences of different types of 'interneurons' to form synapses on specific domains of nearby neurons. Therefore, the idea of an intermediary neuron already exists.

The paper [21] also quotes [23], which investigates differential signalling, based on selective synaptic modifications. They attribute signal changes to changes in frequency and compare synaptic depression, when it becomes unavailable for use, to the neuron collapse. Differential values can be used to support the idea of analogue values and pressure, and the paper in general shows some support for the ReN ideas. It notes the blocking of channels, that synapses between two neurons can behave differently; and there is some confusion why some block while others facilitate, over what appears to be the same type of neuron.

The paper [29] describes that more than one type of sodium channel can be created and that they interact with each other, producing a variable signal. Small currents are involved, even for Ion channels and they work at different potentials, etc. It is also described how neurons can change states and start firing at different rates. The paper [19] describes that there are both positive and negative regulators. The positive regulators can give rise to the growth of new synaptic connections and this can also form memories. There are also memory suppressors, to ensure that only salient features are learned. Long-term memory endures by virtue of the growth of new synaptic connections, a structural change that parallels the duration of the behavioural memory. As the memory fades, the connections retract over time. The paper [2] argues for a mechanical process, with a pressure loss, which is favourable for this paper was well. The paper argues that an Ion pump is not sufficient to explain the forces involved; where one might think that fluid pressure plays a part in the construction of the synapse and also in keeping pathways open, for example. It also argues that a steady pressure state must be maintained throughout the brain and; as the brain needs to solve differential equations, it needs to be analogue in that respect. To support this might be the paper [31] that describes how brain perfusion is highly sensitive to changes in $CO_2$ or $O_2$. An increase in blood flow and resulting gases would possibly increase pressure as well, where their results also challenge conventional theory.

So, there appears to be constructive synaptic processes and these can form memory structures. Excess signal is definitely produced and must be dealt with. There are different types of neuron and sodium channel, where the sodium channel represents a variable signal. The signal or excitation is normally thought to be a mixture of the Ion channels, more than through pressure from any type of momentum. Intermediate neurons already exist. If there is some evidence for pressure changes, then that is good for this paper. For something the size of a synapse channel however, it would be very small.

## VII. MATCHING WITH THE EARLIER RESEARCH

A number of earlier papers ([6]- [11])have been published that describe different aspects of a cognitive model that might resemble a human brain. There was no particular plan to combine them all, but either by chance or through some common underlying theories, it is possible to put a lot of the previous research together, into a single coherent model. This would be a model that would operate in a way similar to what a real human brain does, but might not model it exactly. Earlier research started with [11] or [12] that looked at how a generic cognitive model might be developed from very simple mechanisms, such as stigmergic or dynamic links. These links are created through dynamic localised feedback only and therefore do not rely on any particular rule or piece of knowledge. They are therefore much more flexible, as the feedback would be able to represent anything. The construction process is also largely automatic or mechanical, which is a key requirement. The following sections describe the other related models.

### A. Symbolic and Arbitrary

The model of [10] suggested a new neural network construction that is symbolic in nature. While not important to the ReN, this is seen as an advantage in some cases, because the network can then be understood externally and reasoned over. Each node can represent an individual concept and clustering can be determined locally, between only a few nodes. As well as the symbolic (labelled) concepts however, the network can also create intermediary unlabelled ones that only it will understand. Clustering is based on time, where events presented at the same time get clustered into a unique hidden-layer group. Its construction process is dynamic, where the connections will continually change and nodes are added or removed. Figure 3 gives an example of what it might look like, where the middle layer clusters low-level concepts into higher-level ones. The network is trained in one direction to cluster, but used in the opposite direction to retrieve feature sets. It might then be the global concept that gets asked for, to return the individual parts. For example, asking for the food items (C0 – C4) of a recipe (GC0). Each intermediary group is a separate instance, which allows it to filter out noisy values better. As part of the learning process however, each weight update is then very specific. For smaller datasets this is not a problem, but for the real world it probably would be. This has to be balanced against an alternative of forming a single cluster if the input node sets overlap, where t0, t1 and t2 would form a single group, but cannot then be told apart.

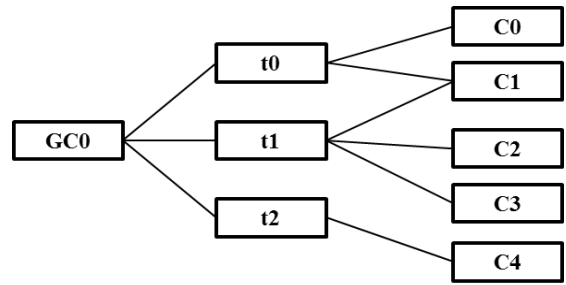

Figure 3. Schematic of a three layer symbolic neural network.

For more realistic data, real-world images for example, might require a more fuzzy form of feedback reinforcement. Lots of intermediary units will be created and destroyed, so for any to survive might require some flexibility on the edges of a cluster, so that the core feature can have slightly different variations at the edges. An alternative to this would be to make the 'feedback' a bit fuzzy instead and keep the hidden layer nodes as they are. So each unit remains distinct, but the feedback is to an area, not a specific unit, as in the real brain. Therefore, separate clusters that are 'conceptually' grouped inside of the selected area can receive common feedback and be updated. As a larger cluster group is rarer because of consistency, it makes sense to use nesting this way. In Figure 3, for example, if the group C1, C2, C3 is presented again (t1), a group consisting of C2 and C3 could also be reinforced. One consisting of C1 and C4 would not. This is also interesting because the paper [6] argues for a nested structure for representing concepts.

### B. Concept Trees

Another paper [7] suggested a way of linking concepts or knowledge that uses a very simple rule that would allow them to be created automatically and be generally applicable. By only allowing concepts that have a lower frequency count value, to be linked to concepts that have higher or equal count values, a sound structure can be generated. If some concept at a tree branch realised a higher count value than its parent, the branch would be broken off to form the base of a new tree. This has a normalising effect on the structure, where the nodes that are used more often, are placed at the tree bases, from where they can be indexed. An example of this is shown in Figure 4, which is similar to the paper diagrams.

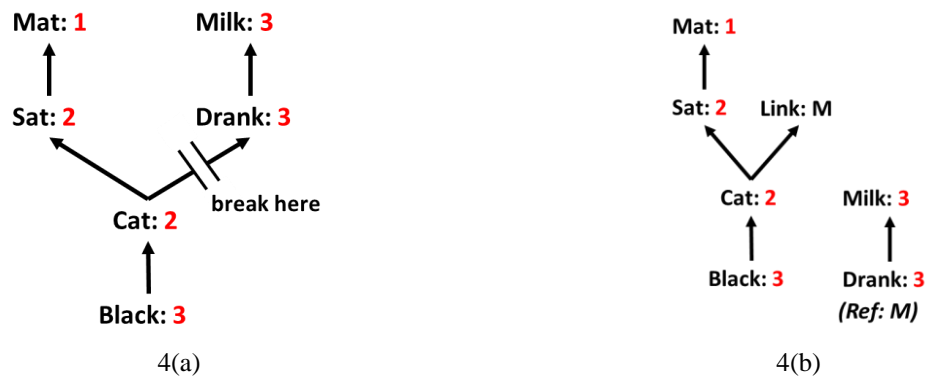

4(a)  4(b)

Figure 4. Example of a Concept Tree, with a new base being created.

Text has been clustered to produce 'black cat sat mat' and 'black cat drank milk' tree paths. In figure (4a), the 'drank – milk' branch has been updated more times than the 'black – cat' root, suggesting that it might be better as the base of a new tree. This is therefore automatically performed, resulting in two trees, as shown in figure (4b). A dynamic link between the two trees (*M*) is then added, to maintain the path. These 'concept trees' are also logical with respect to Nature in general and the intention is to research this more in the future. Tree structures themselves naturally tend towards leaf nodes at the end of terminal branches and it is a bit like the terminals of Minsky's frames [24]. The search is bottom-up, but the links maintain integrity. Unless 'black cat' is asked for, the tree will not be searched, but once started, if there is no 'M' branch link, the search cannot go from tree 1 to tree 2.

If trying to use this design as part of a brain structure, then some ideas that are not fully formulated could be the following: Nodes are linked by synapses, where groups can differ from each other. It is also known that some sort of hierarchy exists (for example, [14]). The idea however is that through their natural formation and use, ensembles with many connections and possibly larger collective sizes, will link to other groups with fewer connections and smaller sizes, until the end groups have almost no extensions and are particular in some way. These end points represent final ideas/objects/concepts that the brain knows about and would not need to investigate further. If thinking about a coffee cup, for example, the brain does not try to search further into molecular structures, but stops at the coffee cup concept (usually). If the connection structure naturally tends to smaller and smaller entities, then the search can stop when it hits one of the more singular or end ensembles. This would have to fit in with the traditional theories of completing circuits, but as other theories also note terminal points that should be possible. The dead end is then the point from where the resonance can start. As suggested in section VIII however, this might require re-combining concept tree leaf nodes into more singular units again.

### C.     Memory and Search

A search to retrieve a piece of memory or one to retrieve a process can maybe behave differently. Even if they both start at the concept tree base, retrieving a memory request can end at a concept tree leaf, for example. A justification for this can simply be that there is not enough support from a memory request to expand the search into the dynamic processes area, or the neural network of Figure 6. The idea of concept trees, or linking through sub or nested patterns, fits nicely with a memory structure and can be built largely from existing static knowledge. A memory recall is usually quite quick and usually for more simplistic pieces of information. If the design puts these at the end of tree structures, memories are these smaller trees that terminate. A more complex problem might require more than one initial concept to fire and for their searches to combine or agree over a wider range of paths. They would start searching from their bases for associations to the more complex problem and might need to combine to reinforce certain channels or ideas. These combined searches then also need to find common terminating conditions. This more dynamic process is considered again in the following sections. So a concept tree is a static piece of knowledge that gets searched over. It can be created and changed dynamically over time, but that would be a slower process. Searches are very quick, but can only search over the already created knowledge and therefore may need to cross-reference paths, to generate a rich enough pool to start with. Different indexing can be used however to access the initial base tree concepts, or to traverse between trees, providing added levels of dynamism. So the search possibly expands first, before contracting to the terminal states.

One question is: can a search only ever stop at a physical world object of some sort? This is because the terminal state must be recognised and so must exist in some form. Concept trees can accommodate this, where the base concepts would be the physical objects that then link to different scenarios at the leaf nodes. The scenarios are specific instances or descriptive examples, possibly with specific values sets. For example, a red shirt. Memory recall might index single base concepts, but a search to solve a problem can find sets of basic objects, apply constraints to find specific instance examples and also allow these to traverse across each other for reinforcement. It is also appealing if thinking about very young humans or babies. They probably learn to recognise the physical objects first and then later start to form associations and processes between them. So the natural process would be to create the physical objects first. Another question that could be asked is: can anything really new be thought? For this paper, the question is more about 'thinking' of something new and not reasoning over a new real-world situation. If searches trace over existing structure, then it might require changing the structure, which might be more difficult without real-world references to index. That is possibly what invention is, which is also why it is so difficult. The paper [13] has also asked this question, with respect to the ART model and also provided some technical answers. In that case, it appears to deal more with a real but unfamiliar world, than a person's own internal imagination.

### D. Aggregations and Stimulus

The papers [6] or [8] and this one describe how stimulus is key to the thinking process and how intelligence might be the result of an ensemble or aggregation of a large number of neurons, all firing together. While this is already known, if it is the case, then that makes it easier to replace one concept in a thought process by another one. We sometimes do this when reasoning, to create new scenarios that never actually existed, but still contain our known elements. It is also a well-known memory technique. The replacement of part of a thought would be easier if all of the neurons fired together in an ensemble. It is difficult to imagine that the brain would surgically move specific concepts as part of the search process. It is more likely to be the case that different neural ensembles would overlap and balance themselves against desired responses. For example, consider listening to recorded music. Each segment of the recorded track is a single noise, but when played in sequence, we can tell each instrument from the other. The brain can separate out from the aggregated sound, the different musical instruments, as the pattern changes.

### E. Resonance and Pressure

This is possibly the most controversial or unlikely part of the new model in a biological sense, but uses the idea is that neurons in-sync with each other do the same thing. The papers [2] or [5] might also have had similar ideas. For this process to work, it is required that a signal travels in both directions through the synaptic channels. It might be more commonly thought that information flows in a one-way system, but cycles back to complete circuits for reinforcement purposes. Cycling is necessary to sustain the signal and also possibly for history recall, but a direct feedback has other advantages. Consider the experiment for creating a standing wave from a moving rope. It is achieved by two people holding the rope and moving it quickly, up and down, to send waves in opposite directions that collide and combine, to produce the standing wave. So this is the proposed new mechanism for the creation of resonance that the brain signal creates. When a dead end (firing blocked or singular concept) is found, this sends the signal back again, where it combines with the signal flowing in the other direction to produce resonance. The whole search path can be recognised as it gets excited by the feedback. It is almost easier mechanically for the path already created, as part of the search, to receive feedback that tells specific neurons that they are relevant. Figure 5 tries to show these new ideas in action, as part of a search process. This is only one part of a larger search group, is very schematic and is simply trying to illustrate an idea. As the forward and backward signals collide and create turbulence, the whole brain area can get excited and resonate. How a neuron would close or become a terminal one however, is not clear.

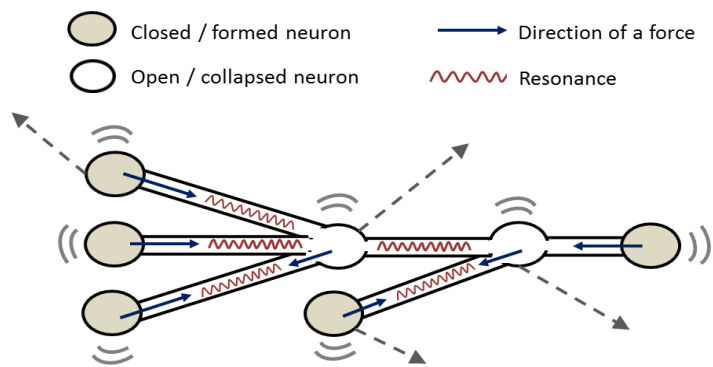

Figure 5. The new neuronal model in action.

The complete cognitive model that is being considered is still the original model, described in [8] or [11], for example. It consists of 3 levels: The bottom level is mostly for optimising and creates individual links between nodes. The middle level is for aggregating the bottom level links. The top level is then for clustering the aggregated sets of links, for higher-level reasoning. The middle level's self-organising abilities would be useful and different neural network models could be tried there, including the new ReN structure that is suggested in this paper. The top level can make use of concept trees [7], or the neural network model of [10], for example. It might be best to view the different levels as functional and not physical. The linking operations are different, but they can all occupy the same physical space and work on the same single structure.

### F. Regional and Temporal Feedback

Time is maybe slightly ignored in computer models based on neural networks, but essential for any state-based system [3]. Recognising specific areas or regions is also a problem to be solved. The symbolic neural network of section A works well, as it can filter out noisy input, by using very specific clusters. When training however, the reinforcement updates need to be equally exact. As suggested, a more general structure to the feedback mechanism could be tried. The papers [14] and [21] note the requirement of a temporal feedback, so that historical events are recognised and can then be predicted or reasoned about. Hebb's rule would suggest that similar groups would form closely to each other, as opposed to far apart, if no other influence is involved. Therefore, instead of the feedback going to one specific cluster set, let it go to a general area and influence all of the relevant cluster groups in that area. A detailed discussion of that is too much for this paper, but it is included to try to keep in-sync with a real biological model. The point is that similar concepts are likely to be grouped closely together and so it makes more sense that a general feedback could influence more than one of them at a time. Think of a line of patterns or states that trigger each other. If the correct result is state 1, then if we also reinforce state 0 or state 2 that is still probably OK.

## VIII. ADDING DETAIL TO THE COGNITIVE MODEL

On writing this paper, it has become clear that the neural network and the concept trees can be closely related with each other, both in terms of their structure and also how they are formed and used. While the concept trees are more static and can be built from existing knowledge, the symbolic neural network is more dynamic and influenced by time. A search process might start with specific concepts, broaden out and then narrow to terminating states. A search is also involved for real-world problem-solving or predictions and not just for direct memory retrieval. So another idea of this paper is to join the neural network with the concept tree, to produce a more autonomous system. It is autonomous because full processes are now possible, from the automatic ones previously described. It can be searched over at different levels, store existing knowledge, or produce previously unknown concepts from experience. This is illustrated in Figure 6, where similar types of architecture have probably been proposed before, for example [4], [5] or [13]. To complete the picture, the indexing to the base of the concept trees is provided by a lower level of stimulus. This can be refined using the ReN neurons, for example that act as the indexers to the structured knowledge in the higher level. Each part of the architecture has a construction stage and a usage stage. Time is again key, where node groups presented during the construction stage do not have to be the same as the groups presented during the usage stage. A terminal state might be a conclusion to a process, as in an act and not a physical object, but it is still something that has existed previously.

In the top level, one can see how the concept trees, built from existing knowledge can provide a platform on which the neural network can be built. If a set of tree nodes are used as part of a particular search result, then that is an abstract time-based group that represents something. If the cognitive model wishes to store that information, it could use the neural network to cluster it into the hidden layer units and then the single global concept. So this could marry the static knowledge with a more dynamic reasoning process, but how it is accessed or searched over is still unclear. A memory-based search might only use the concept trees, by presenting a set of indexes to the base tree nodes only. Remember that a search can also link between trees. A process search might then require a circuit that connects with the global neural network concepts as well. If the C0 to C4 concepts of Figure 3 were replaced by more descriptive concept trees, then the clustering principle is the same, even if the whole process is more complex.

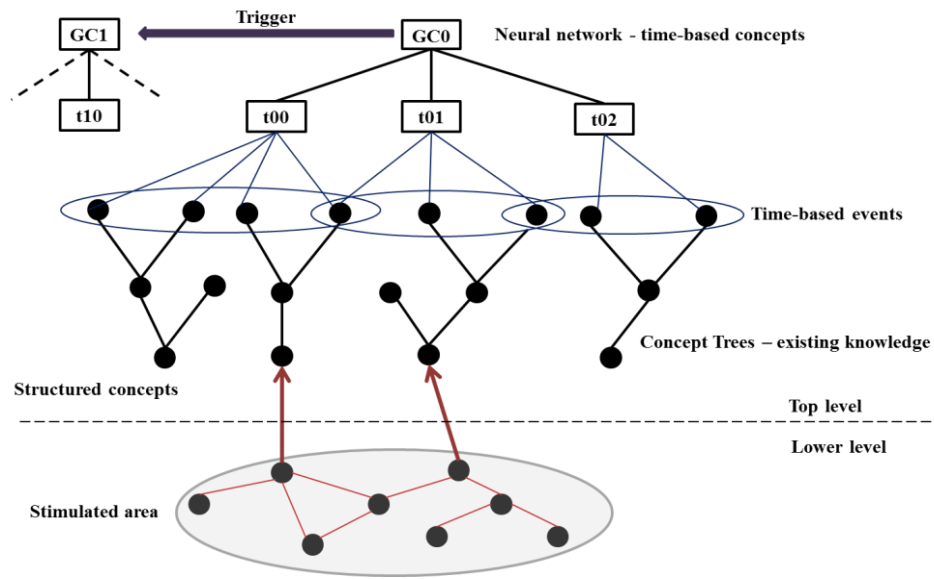

Figure 6. Integration with the Cognitive Model of [8][11], built from the new clustering structures.

The original cognitive model, described in [8] or [11], for example, already has an upper-level structure for higher-level concepts that can trigger each other. The triggering acts as a form of process creation and allows for more intelligent thought. Integrating the new model components, it is possible to simply allow triggers between the global neural network concepts, as they already represent more complex entities and the sub-entities have not been defined exactly. As a hierarchy is already known about, the lateral connections are just another feature of it. For the concept tree / neural network combination, the same architecture is described in [7], section 7.3. With these two construction processes, one relates to knowledge-based concept trees and to small but specific entities, while the other relates to the event-based clustering and also to self-organising these smaller structures. Following earlier papers' food examples, you could imagine a human tasting different food types and learning what they are; but when in a restaurant, selecting a menu item based on the food types and discovering some new recipe through the experience. The lower level stimulated area is added to try to complete the picture. The exact functionality there is expanded on in other papers, but is more aggregating in nature. The idea of preferred or more significant neurons is not impossible. As described in section V.B, a neuron that fires more often might be able to keep its direct path. If it is then involved in a firing pattern, it will have more influence, suggesting that it has a more significant value. A stimulus input could even be thought of as the query request, leading to the activation of concept tree bases.

## IX.    DISCUSSION

Looking at a cross-section, or magnified image of a real brain, shows how completely random the structure is. A computer model would struggle to create this sort of structure and prefers a much more ordered one. While the new model is mostly concerned with the search and retrieval process, the first part of this paper showed an arbitrary construction process as well, showing how the channels that get searched over might form in the first place. The resulting simplified neuronal operation would give added confidence that the same type of neuron can be modelled reliably, including threshold and output signal strength. It can produce a desirable variety of signal strengths through a layered hierarchical structure that can be created as part of any random process. Although, its usefulness needs to be tested further. As far as the brain is concerned, when the appropriate stimulus is encountered, all of the neurons that are related to it can fire together, representing the same concept in some way. It could lead to better models of how neurons interact, if more sophisticated sets of values can be considered. Maybe some sensors or input/output connectors with the real world could be connected with the computer simulator to test the theory. The middle or intermediary units are useful for different purposes. If a neuron has a large number of direct inputs, much larger than the required threshold, they would lose some of their meaning. If these are clustered by intermediary units first, then each intermediary unit must represent some sort of abstract concept by itself, even if it is not a real world one. The excess feedback would almost force the creation of these sub-concepts, as part of a self-organising system. If pressure plays a part, then excess amounts would need to be adjusted. There is also the original point of the theory - to produce that apparent analogue set of values, from binary-operating neurons. This type of model is actually less complicated than the real one, but it has the problem of passing a signal in both directions.

    Lots of the features that have been mentioned might be found in other models or technologies, but they probably do not appear in exactly the same format as described in these research papers. While the narrowing tree structure of concept trees is not new, the idea that it can be used as a sort of rule for creating linked structures probably is, although, Markov models may also have the same statistical property. The neural network design [10] is possibly new. The construction and usage directions are largely to do with what it represents, where different cluster levels are involved. It might be a question of granularity as to whether other models inherently contain this process as well however. Aggregations and stimulus is commonly accepted, but the idea that a more refined neuron can be automatically and reliably created as part of that might be new. Memory structures are obvious, but the idea of a terminating condition that results in a signal being fed back down the same channel again, appears to be quite radical. The commonly accepted view is one of cycling and completing one-way circuits, because the signal only travels one way through a neuron. While cycling is required, to reinforce the process, there must be some sort of terminating condition, to indicate a successful result. A resonance feedback process can work in harmony with the other established theories, mathematically at least. It is slightly more forceful than the other processes, but also provides slightly different functionality. Circuits, for example, could be more for search and information retrieval [7], while inhibitors can help with timing or process control [6] and so these established methods can work in the same model as the new ideas, without contradiction.

    The idea of pressure or force is the most radical. It might be the current thought that the brain is made up more of a kind of amorphous gel that can get electrically charged and then change. The resonance option is that the synapse grows into the gel, but it would be an internal pushing force that controls the process. While the new model might, or might not, be 100% correct biologically, it appears to be mathematically sound. This is important, because it is then possible to build a consistent computer model from it that will do something, even if it does not produce intelligent thought, which is the ultimate goal. As the model simulates the real human brain closely, if it is shown to work in some way, then this will probably have relevance to future research in that area. So there is much here that could be used to build an artificial model of the human brain.

## ACKNOWLEDGEMENTS

    This paper is a much extended version of the first version 'The Re(de)fined Neuron', published on Scribd[2](Feb 2013) and also on arXiv[3]. This version has made some important changes or updates and also added a substantial amount of new information, regarding further supporting arguments and a more complete system. It was originally accepted in 2014 for the IGI JITR special issue on: 'From Natural Computing to Self-organizing Intelligent Complex Systems', which was subsequently cancelled.

---

[2] http://www.scribd.com/doc/123586190/The-Re-de-fined-Neuron.
[3] http://arxiv.org/abs/1403.1080.